\documentclass{article}

\usepackage{arxiv}

\usepackage[utf8]{inputenc} % allow utf-8 input
\usepackage[T1]{fontenc}    % use 8-bit T1 fonts
\usepackage{hyperref}       % hyperlinks
\usepackage{url}            % simple URL typesetting
\usepackage{booktabs}       % professional-quality tables
\usepackage{amsfonts}       % blackboard math symbols
\usepackage{nicefrac}       % compact symbols for 1/2, etc.
\usepackage{microtype}      % microtypography
\usepackage{lipsum}
\usepackage{graphicx}
\graphicspath{ {./images/} }
\usepackage{graphicx}
\usepackage{url}
\usepackage{xcolor}
\usepackage{hyperref}
\usepackage{booktabs}
\usepackage{multirow}
\usepackage{tabularray}
\usepackage{adjustbox}
\usepackage{caption}
\usepackage{subcaption}
\usepackage{float}

\title{Knowledge graphs for empirical concept retrieval}

\author{
Lenka Tětková \\ \texttt{lenhy@dtu.dk} \And
Teresa Karen Scheidt \\ \texttt{tksc@dtu.dk} \And
Maria Mandrup Fogh \AND
Ellen Marie Gaunby Jørgensen \And
Finn Årup Nielsen \\ \texttt{faan@dtu.dk} \And
Lars Kai Hansen \\ \texttt{lkai@dtu.dk} \\ \\
Section for Cognitive Systems, DTU Compute, \\
Technical University of Denmark,
2800 Kongens Lyngby, Denmark\\
}

\begin{document}
\maketitle
\let\svthefootnote\thefootnote
\newcommand\freefootnote[1]{%
  \let\thefootnote\relax%
  \footnotetext{#1}%
  \let\thefootnote\svthefootnote%
}
\freefootnote{Preprint. Submitted to The 2nd World Conference on eXplainable Artificial Intelligence.}
\begin{abstract}
Concept-based explainable AI is promising as a tool to improve the understanding of complex models at the premises of a given user, viz.\ as a tool for personalized explainability. An important class of concept-based explainability methods is constructed with empirically defined concepts, indirectly defined through a set of positive and negative examples, as in the TCAV approach (Kim et al., 2018). 
While it is appealing to the user to avoid formal definitions of concepts and their operationalization, it can be challenging to establish relevant concept datasets. Here, we address this challenge using general knowledge graphs (such as, e.g., Wikidata or WordNet) for comprehensive concept definition and present a workflow for user-driven data collection in both text and image domains. The concepts derived from knowledge graphs are defined interactively, providing an opportunity for personalization and ensuring that the concepts reflect the user's intentions. We test the retrieved concept datasets on two concept-based explainability methods, namely concept activation vectors (CAVs) and concept activation regions (CARs) (Crabbe and van der Schaar, 2022).
We show that CAVs and CARs based on these empirical concept datasets provide robust and accurate explanations. Importantly, we also find good alignment between the models' representations of concepts and the structure of knowledge graphs, i.e., human representations. This supports our conclusion that knowledge graph-based concepts are relevant for XAI. 
\end{abstract}

\section{Introduction}
As the range of AI applications expands to all areas of society, it becomes urgent to secure users' understanding and control of AI based systems. Unfortunately, tools and principled approaches for explainability and personalization appear to develop at a slower pace than AI applications. 
While it is well-established that efficient human-human communication is founded in aligned conceptual representations \cite{gardenfors2014geometry}, the role of communication and alignment in human-AI interaction is less explored \cite{christian2020alignment}.

\begin{figure}[t]
    \centering
    \includegraphics[width=\textwidth]{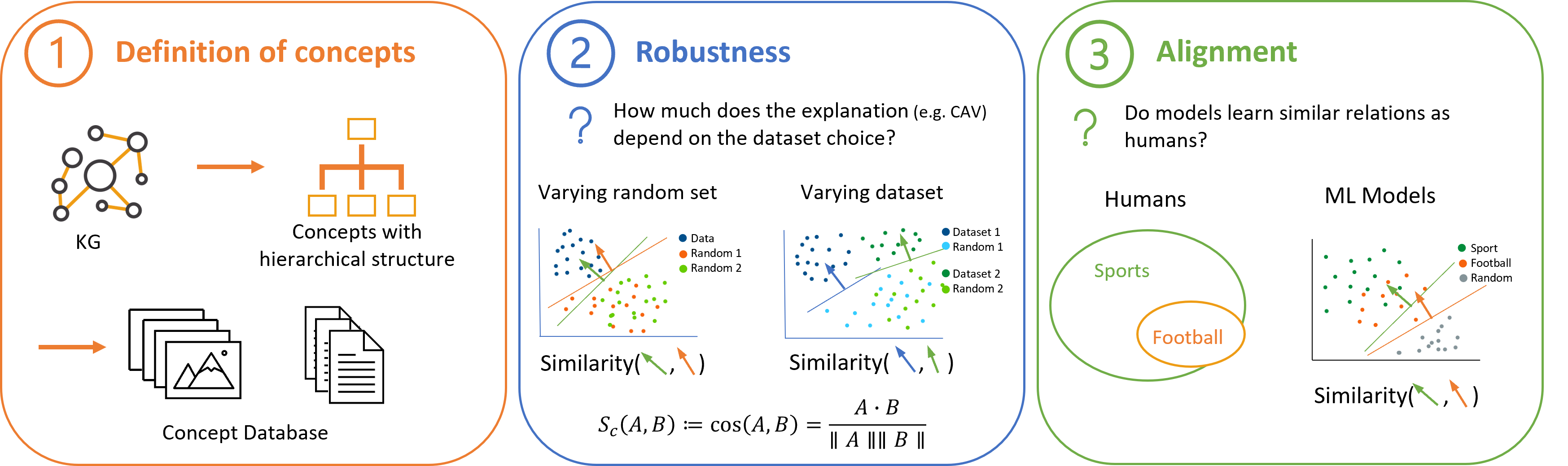}
    \caption{Main research questions and workflow: 
    \textbf{1)} Can knowledge graphs be used to generate data-driven concepts? 
    \textbf{2)} How robust are the explanations (here CAVs) provided by such concepts to variations in the dataset?
    \textbf{3)} Are internal representations aligned with human knowledge? }
    \label{fig:overview}
\end{figure}

The field of concept-based explainability can potentially help to solve the human-machine alignment problem \cite{sucholutsky2024alignment}. An important group of algorithms is based on empirically defined concepts, see e.g., \cite{kim2018tcav}. The assumption here is that they can define a concept empirically by providing both positive and negative examples. This clearly limits the methods to cases based on the availability of relevant data for meaningful concepts. Although formal routes to establish and validate such concept data have a long history \cite{goguen2005concept}, it is largely an unsolved problem. In this work, we propose to use formal knowledge bases as a source of empirically defined concepts, e.g., based on knowledge graphs such as WordNet and Wikidata \cite{ji2021survey}. Knowledge graphs can help to define and delineate concepts, and, as we will show, help to retrieve data for alignment of concepts and machine representations. This approach can generalize existing data-driven concept-based explainability methods that are typically limited to a few datasets with a small number of concepts \cite{kim2018tcav}. In addition, it could help personalize concept definitions by establishing knowledge graph-human interaction to avoid creating a complete personalized concept dataset from scratch, which could be both expensive and time-consuming. Thus, we ask the following fundamental questions: 
\begin{itemize}
    \item Can formal knowledge graphs be used to generate data-driven concepts for explainable AI? - this question addresses the {\it existence} of our approach.
    \item How robust are such data-driven concepts to the choice of modeling approach, knowledge base, and data sampling? - this question addresses the {\it uniqueness} of our approach.
    \item Are the model's internal representations well-aligned with human categorization? - this question addresses the {\it relevance} of our approach.
\end{itemize}
To investigate these questions, we develop several workflows, including tools to enable broad and general concept definition through knowledge graphs, as well as tools for evaluation of robustness and alignment of human and machine-defined concepts. \autoref{fig:overview} gives an overview of the main questions we tackle in this paper. 

With this work, we aim to contribute to several open problems that the concept-based explainability community is facing -- namely, the need for a way to define relevant concepts (preferably in a personalized, interactive, and hierarchical way to allow adaption to specific needs), the need for robustness of XAI methods, and the problem of human-machine alignment \cite{logo2024XAI-manifesto}.
Our specific contributions include:\\[-6mm]
\begin{itemize}
\itemsep 0pt
\parsep 0pt
    \item Introduction of knowledge graphs as a tool for concept definition in concept-based XAI.
    \item An interactive approach to concept definition and retrieval.
    \item Approach for evaluating the robustness of concept activation vectors and regions.
    \item Testing the alignment of concept representations in state-of-the-art AI models and human-curated knowledge graphs. 
\end{itemize}
Our results indicate that knowledge graphs can directly be used for interactive concept definition and highlight the great potential of Wikimedia resources for empirical concept data retrieval. We find concept activation vectors/regions based on datasets defined with our approach are robust and accurate, while providing the possibility to align the concept definition well with user's intentions. Lastly, we demonstrate that concepts and related sub-concepts (defined by curated knowledge graphs, i.e., human cognition) share similar concept activation vectors/regions, which is direct evidence of human-machine alignment.

\section{Related Work}
\subsection{Concept-based Explainability Methods}
Concept-based explainability has been a topic of interest in recent years, providing human-understandable explanations via higher-level attributes, usually referred to as concepts. Various theoretical frameworks have been proposed in recent years, most distinctively post-hoc methods and inherently interpretable methods. For a detailed overview of all concept-based methods, the reader is referred to \cite{poeta2023conceptsurvey}, here we mention only the most relevant contributions. 

Post-hoc methods aim to provide explanations after a model has been trained. A significant contribution in this area is the work by Kim et al. \cite{kim2018tcav} introducing Concept Activation Vectors (CAVs). CAVs are defined by training linear probes to separate positive and negative samples of a given concept. The inner product of the directional derivative of a classifier and the CAV is a quantitative measure of the classification model's alignment with a given concept. Several methods have explored and generalized the idea of CAVs, noteworthy is the generalization to concepts activation regions \cite{crabbe2022car} relaxing the linear separability assumption. Other post-hoc methods include Interpretable Basis Decomposition for Visual Explanation (IBD) \cite{zhou2018IBD} and Causal Concept Effect (CaCE) \cite{goyal2019cace}. The latter investigates the causal effects of the presence/absence of concepts. Several open problems remain with these methods, especially concerning the concept data. For example, the explanations can be sensitive to which concept data set is invoked and CAVs may be less well-defined than the classes they are thought to explain (less accurate) \cite{ramaswamy2023overlooked}. 

While all of the methods mentioned above require pre-defined concepts with examples, there are also approaches aiming to discover the underlying concepts that a model has learned. The approach to discovering and defining concepts varies between methods. In Automatic Concept-based Explanations (ACE) \cite{ghorbani2019ace}, relevant concepts are extracted by segmenting all class images and clustering of similar segments, which will then be used to calculate CAVs. Concepts are here defined as different-sized segments, covering coarse features (e.g., full objects) and fine features (e.g., textures, colors). Invertible Concept-based Explanations (ICE) \cite{zhang2021ice} builds on ACE, but uses a CNN as a feature extractor to generate examples for concepts, and matrix factorization is applied to provide CAVs. Concept Recursive Activation Factorization for Explainability (CRAFT) \cite{fel2023craft} aims at combining classical XAI methods based on feature attribution with concept-based explanations. Here, examples of concepts are defined as random crops of images. Activations from these examples are decomposed using non-negative matrix factorization, forming a concept bank. Adding recursivity over multiple layers, they build hierarchies of concepts and corresponding sub-concepts. In Multidimensional Concept Discovery (MCD) \cite{vielhaben2023multi}, concepts are discovered completely without predefinition of concepts by decomposition of the feature space. A concept here is defined as a multi-dimensional linear subspace in the feature space.  

So-called ante-hoc methods incorporate interpretability into the model architecture itself. One well-studied example is the family of Concept Bottleneck Models (CBM) \cite{koh2020cbm}, where concepts are predefined with annotated examples to train bottleneck layers, making subsequent classifications interpretable and intervenable. Post-hoc concept bottleneck models \cite{yuksekgonul2022postCBM} (PCBMs) and label-free concept bottleneck models \cite{oikarinen2023labelfreeCBM} (label-free CBMs) are two approaches to make any already trained model a CBM by adding an interpretable bottleneck layer. Both models introduce the use of multimodal models (CLIP \cite{radford2021clip}) to alleviate the need for labeled concept data. PCBMs use ConceptNet \cite{speer2017conceptnet} to define relevant concepts, while label-free CBM uses GPT3 \cite{brown2020gpt3} to define concepts. Both methods then use the text embeddings of CLIP as their concept base to train the bottleneck layer. Our approach is similar to these examples of concept definition, but we are adopting a broader approach to explore the unique capabilities of KGs in the explainability context. This includes the careful curation of the used KG resources, stability, and reproducibility that may not be found in applications based on Chatbots, such as GPT3. Several other works \cite{Kim2023pcbm,Marconato2022Glancenets,sawada2022cbm,zarlenga2022cem} have built on CBMs, tackling various shortcomings of CBMs, such as data leakage and reduced accuracy. 

While several methods are developed that do not rely on predefined concepts or annotated examples of concepts, extant work depends on some form of empirical concept definition and thus requires examples or annotated concept data. There exist a few data bases that are often used for defining concept data, such as Pascal VOC \cite{pascal-voc-2012}, Broden \cite{bau2017broden}, ADE20k \cite{zhou2017ade20k}, MS COCO \cite{lin2014mscoco} and CUB \cite{wah2011cub}. However, these datasets only cover a limited range of all potential concepts. More specific concept data can be challenging and costly to obtain, especially for domains where concepts are ambiguous, subjective, or evolving. Not much work has focused on how to obtain a fitting concept database. This motivates the current work, where we take a knowledge graph-based approach to define relevant concepts and automatically retrieve examples. 

\subsection{What is a concept?}
Until now, we have appealed to an intuitive notion of concept.
While there are many concept-based methods developed, most use the term quite vaguely. Formalizations exist in cognitive science and the more mathematical literature, e.g., \cite{fauconnier1994mental,faueonnier2002way,gardenfors2014geometry,goguen2005concept,gardenfors2000conceptual}.
However, these theories are not directly aimed at practical/computational concept representations, hence, can be hard to use in practice. Poeta et al.\ in \cite{poeta2023conceptsurvey} categorize concepts as they are used in XAI into four categories: symbolic concepts, unsupervised concept bases, prototypes, and textual concepts.
In the present work, we assume symbolic concepts derived as human-defined symbols through empirical workflows requiring auxiliary data equipped with implicit concept annotation as produced by knowledge graphs.

\subsection{Knowledge graphs}
Knowledge graphs (KGs) are structured representations of semantic information, which can hierarchically represent knowledge. KGs are composed of nodes representing entities (e.g., people, places, concepts) and edges representing relationships or properties between these entities. There are several KGs freely available \cite{tiddi2022KGsurvey}, with different characteristics such as different scopes, sizes, and sources. We establish workflows for three well-known KGs, namely WordNet \cite{miller1995wordnet}, Wikidata \cite{vrandevcic2014wikidata}, and ConceptNet \cite{speer2017conceptnet}, as these are widely used and allow for easy interaction. WordNet is a lexical database, linking words by their semantic meaning. Wikidata is an open and collaborative knowledge base that acts as a central structured storage for all Wikimedia projects. ConceptNet is a freely available semantic network that captures the meaning and common-sense knowledge of words. 

KGs have been used to define concepts for concept bottleneck model \cite{oikarinen2023labelfreeCBM,yuksekgonul2022postCBM}, to instill prior knowledge into models and to improve explainability \cite{tiddi2022KGsurvey}. Some problems have been pointed out, such as the need for quality control and filtering of relevant information \cite{icarte2017general}, the problem of identity alignment between KGs, and how to automatically extract knowledge \cite{tiddi2022KGsurvey}.
KGs have been noted to be relevant for concept-based XAI \cite{lecue2020roleKGinXAI,logo2024XAI-manifesto}, yet practical approaches to empirical concept definition, as in this work, have not been proposed.

\section{Methods}
We provide an interactive workflow for defining concepts based on KGs (\ref{sec:KG}), and examples for automatic concept data retrieval based on Wikimedia Commons and Wikipedia (\ref{sec:data}). We test the impact of the concept database on the activation vectors and regions (\ref{sec:acc_robustness}) and investigate whether the concepts and sub-concepts defined by humans are aligned in the internal representations of the models (\ref{sec:alignment}). The code is available on GitHub\footnote{\url{https://github.com/LenkaTetkova/Knowledge-Graphs-for-Empirical-Concept-Retrieval.git}}. 

\subsection{Concepts from knowledge graphs}
\label{sec:KG}
The first step in defining a concept database is to formulate meaningful concepts. We propose to let the formulation be guided by KGs, as they offer rich resources for capturing information and relationships among entities. We describe the process for three commonly used KGs: WordNet \cite{miller1995wordnet}, Wikidata \cite{vrandevcic2014wikidata}, and ConceptNet \cite{speer2017conceptnet}. The suggested workflow is easily adaptable to other KGs, such as other openly available KGs, but also to specialized KGs for, e.g., medical use cases. 

\subsubsection{WordNet}
WordNet is a lexical database, which captures linguistic relations between words. It relies on a hypernym-hyponym relation, which indicates the type-of or kind-of relation between concepts (so-called synsets in WordNet). It covers rich semantic relations and is curated by experts (linguists and lexicographers) who ensure that the synsets and relations are accurate, consistent, and coherent. WordNet also has a well-defined and documented methodology for building and maintaining its lexical database and follows established standards and conventions for its data format and representation, making it very reliable and easy to work with. WordNet contains over 100.000 synsets. ImageNet \cite{ImageNet} is built on the WordNet hierarchy and provides a great resource of images related to WordNet entries. Note, however, that the synsets of WordNet are a superset of the sets represented in ImageNet. 

KGs in WordNet can be easily extracted using the nltk\footnote{\url{https://www.nltk.org/}} toolbox and the hypernym-hyponym relations. We extract two levels downwards (hyponyms) for each general concept to get more specific entities of each concept. For exploration, we also extract hypernyms and their hyponyms for each concept to ensure the concept is as specific or broad as wanted. Once the desired KG is extracted, all nodes in the KG are mapped to Wikidata entries to ensure disambiguation and enable the next step of data extraction. This is done interactively by choosing the most relevant Wikidata item (based on its description) from all suggested items that fit the concept name. This way, we can ensure the related Wikidata entries describe the wanted meaning. 

\subsubsection{Wikidata}
Another frequently used knowledge graph is Wikidata \cite{vrandevcic2014wikidata}, which serves as a factual knowledge graph. Wikidata is linked to Wikipedia, the Wikimedia Commons media archive, and other wikis, making it a valuable source for extracting knowledge and related media such as images, text, and audio. It is composed of items, each identified by a unique identifier, and these items are connected through properties. Wikidata consists of over 100 million items, of which 23,000 are categorized as a class (which we consider a concept), and over 11,000 properties that can link items to each other and external sources. 

The Wikidata Python library\footnote{\url{https://wikidata.readthedocs.io/en/stable/}} or SPARQL queries can be used to directly extract KGs from Wikidata. For this, the wanted properties need to be defined, which can vary greatly depending on the use case. Some general properties include 'subclass of', 'part of', and 'has use'. To build a hierarchical KG, similar to WordNet, we use the property 'subclass of' (P279) and 'instance of' (P31) to and from the queried concept.
We also use Wikidata as a secondary source to ensure the disambiguation of concepts. Using the Wkidata API\footnote{\url{https://www.wikidata.org/w/api.php?}}, we link each extracted concept to a Wikidata identifier, which has a specific description. This way each concept has the desired meaning, e.g., apple (Q89) vs. Apple (Q312).

\subsubsection{ConceptNet}
ConceptNet is a multilingual knowledge graph that connects words and phrases of natural language with labeled edges, representing various semantic relations such as 'IsA', 'UsedFor', 'RelatedTo', etc. ConceptNet focuses on representing common-sense meanings and relations of words. It is based on crowd-sourcing, expert-curated resources, and gamification. ConceptNet contains more entities and relations than WordNet, it contains over 200.000 concepts and over 30 relations \cite{speer2017conceptnet}. 

Concepts and their relations can be extracted using the ConceptNet API\footnote{\url{http://api.conceptnet.io}}. The relations can be adapted for the use case. We use the relations 'IsA', 'MadeOf', 'HasA', 'HasProperty', and 'PartOf'. The extracted related concepts need to be cleaned. We remove duplicates, remove preceding articles (a, an, the), and remove concepts with more than three words. This cleaning can be extended to remove concepts with very similar meanings or less reliable sources as done in \cite{oikarinen2023labelfreeCBM}. 

\subsubsection{Other options}
Another route to define concepts is using LLMs, more specifically GPT-3 as in \cite{oikarinen2023labelfreeCBM}. However, these are not always reliable and do not provide reproducible answers, and LLMs have also been shown to lack common sense and causality \cite{mahowald2023dissociating}. This may reduce trust and is at odds with the main objectives of explainable and trustworthy AI. By using openly available knowledge graphs that are curated by experts, we can provide a framework for extracting reproducible and reliable concept sets. Note that LLMs with integrated KGs may fix some of the current LLM problems and provide a valuable resource for the workflow in the future \cite{pan2024unifyingLLMsKGs}.  

For specialized tasks (e.g., medical use cases), appropriate specialized KGs (e.g., BioKG \cite{walsh2020BioKG} or PharmaKG \cite{zheng2020PharmaKG}) can be invoked to extract domain-relevant concepts and relations. 

\subsubsection{Choice of knowledge graph}
\begin{figure}[ht]
    \centering
    \includegraphics[width=\textwidth]{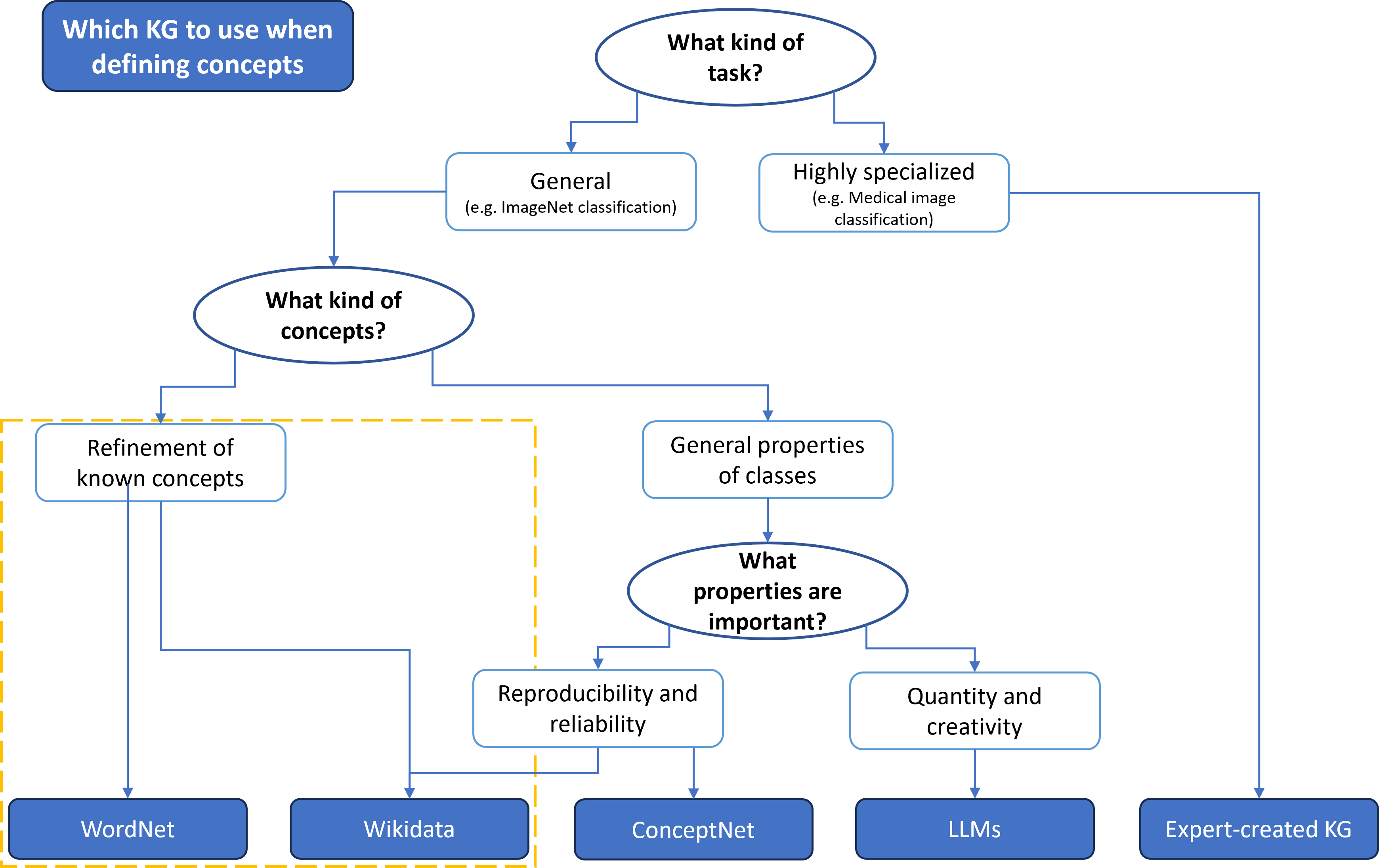}
    \caption{Decision tree for choosing an appropriate knowledge graph based on the individual needs and wants. This is to give a general overview, individual decisions might require more consideration or different KGs. We will use WordNet and Wikidata (indicated by the orange box) for our main experiments.}
    \label{fig:KG_decision}
\end{figure}

The choice of KG depends strongly on the use case. For example, while ConceptNet is a good choice for extracting common-sense properties and relations for known classes, WordNet can be used for a wider exploration utilizing the hierarchical structure to provide different levels of concepts, allowing for customization of concepts. \autoref{fig:KG_decision} provides a general overview that can help guide the decision process. Our experiments will mainly focus on WordNet and Wikidata.
\subsubsection{Interactivity of concept definition}
Advantages of this approach include availability, interactivity, customization, and reliability resulting from using a curated and trustworthy knowledge graph. Especially the interactivity can lead to a well-defined concept that is fitted to the needs and intentions of the user. We created an interactive pipeline to define the concepts:

The first step is the main concept definition, let us illustrate it with an example:
 Imagine that you want to explore the concept of \textit{house}. First, you have to choose among different suggestions to disambiguate your search. The system gives you the following options: ``house - building usually intended for living in", ``House - American television medical drama", ``house music - electronic dance music genre, that originated in Chicago in the early 1980s" and ``United States House of Representatives - the lower house of the United States Congress". After you choose the one you prefer (in this case the first one), you get a set of data examples characterizing this concept. You can review it to see if it encompasses what you had initially in mind. If not, the system suggests other concepts either one level up in the hierarchy (e.g., building) or one level down (e.g., cottage, detached house, or tree house). The user can navigate up or down in the knowledge graph and get concept data until the necessary detail is achieved. The outcome is a set of data points defining application and context-dependent concepts.

The second (optional) step is including relevant sub-concepts in the database. For each chosen main concept, you get a list of sub-concepts that can be included in the search query if wanted. This leads to a more diverse dataset and inclusion of all relevant examples you had in mind. Each sub-concept also needs to be linked to a unique identifier in Wikidata to ensure the right meaning of each concept. This is also done interactively, so each concept fits the intention of the user: For each concept, the user gets a list of all possible definitions and can choose the one that fits best (or none). This way the user has efficient control and can ensure each selected concept (and its data) has the intended meaning.

In the next step, data related to every defined concept is retrieved in an automated way. Due to the disambiguation of concepts, the extracted data need minimal supervision to ensure quality. 

\subsection{Retrieval of a concept database}
\label{sec:data}
Once the concepts and their relevant KGs are extracted and matched with unique identifiers, the concept database can be built. This step is highly dependent on what kind of data is needed and what resources are available. We suggest two workflows, one for images and one for text, to collect data for each concept using Wikimedia Commons and Wikipedia as sources. 
\begin{table}[ht]
  \centering
  \begin{subtable}{\linewidth}
  \centering
  \begin{tabular}{|l|l|l|l|} \hline
Concept name     & WikiData ID   & Pascal VOC    & WikiMedia     \\ \hline
bicycle          & Q11442        & 552   & 3918          \\ \hline
bottle   & Q80228        & 706   & 1590          \\ \hline
cat      & Q146          & 1080          & 31158         \\ \hline
potted plant      & Q27993793     & 527   & 1290          \\ \hline
sheep    & Q7368         & 325   & 3596          \\ \hline

    \end{tabular}
    \caption{Number of images contained in each database. All images from Wikimedia Commons are linked to the Wikidata ID ensuring the right meaning of the concept. The name of the concept corresponds to the label in Pascal VOC.}
    \label{tab:image_database}
  \end{subtable}

  \begin{subtable}{\linewidth} 
  \centering
  \begin{tabular}{|l|l|l|l|} \hline
     Concept name       & Wikidata ID   & \# images     & \# sentences          \\ \hline
random                  & -             & 16225         & 544765              \\ \hline
\textbf{sport }                  & Q349          & 22558         & 609456        \\
gymnastics              & Q43450        & 183           & 530   \\
cycling                 & Q53121        & 450           & 8688          \\
\textbf{fruit}                   & Q3314483      & 1818          & 2266          \\
apple                   & Q89           & 1431          & 313   \\
berry                   & Q13184        & 192           & 119   \\
\textbf{motor vehicle}           & Q1420         & 29653         & 57504         \\ 
tow truck               & Q332050       & 128           & 55    \\
minivan                 & Q223189       & 156           & 157    \\ 
\hline
  \end{tabular}
  \caption{Images from Wikimedia and text from Wikipedia for some \textbf{concepts} and sub-concepts.}
  \label{tab:subconcepts_database}
  \end{subtable}
  
  \caption{Dataset statistics (for a subset of concepts) of the data created from Wikimedia and Wikipedia.
  }
  \label{tab:dataset_statistics}
\end{table}

\subsubsection{Images - Wikimedia Commons}
For retrieval of images, we use Wikimedia Commons. Wikimedia Commons is a media file repository with over 100 million freely available image, sound, video, and text files. It is linked to Wikidata, which provides unique identifiers and enables to retrieve specific media files for defined concepts. Wikimedia, therefore, offers a great resource for building an individual and personalized database without the need for close supervision or the time-consuming collection of data.

To build our database, we search the Wikimedia Commons with Wikimedia Commons Query Service for all images connected to (relation P180) the chosen Wikidata item and all its instances or subclasses (relations P31 and P279*). The images were downloaded in their 640x640 pixel size. We then randomly sample from all these images. We compare the results from this database with results from an already labeled dataset, in this case, %ObjectNet \cite{barbu2019objectnet}
Pascal VOC 2012 \cite{pascal-voc-2012}. We test the similarity to the labeled data on the classes \textit{bicycle, bottle, cat, potted plant} and \textit{sheep}. We sample a negative set for each database by sampling random images from non-related classes. \autoref{tab:image_database} shows how many images for each class are available, Wikimedia provides generally more for each class.

We also retrieve images for three concepts and their sub-concepts, namely \textit{sport} (Q349), \textit{edible fruit} (Q3314483), and \textit{motor vehicle} (Q1420). Some examples and their number of available images are shown in \autoref{tab:subconcepts_database}.

\subsubsection{Text - Wikipedia}
We create text datasets using Wikipedia articles. We split the article corresponding to the concept of interest into individual sentences and cut the outlying sentences shorter than 50 and longer than 500 characters. If it does not contain enough sentences, we randomly choose sentences from all other pages that are subclasses of the concept (relations P31 and P279*). As negative examples, we randomly sample more than $25 000$ articles and randomly choose sentences from them.

We retrieve a text database for the same three concepts and their sub-concepts as for images. The amounts of sentences for some examples are shown in \autoref{tab:subconcepts_database}. \\ 

\noindent Once suitable concept databases are defined, we use the retrieved data to run experiments with two established concept-based explanation methods: Concept activation vectors (CAVs) \cite{kim2018tcav} and concept activation regions (CARs) \cite{crabbe2022car}. 

\subsection{Concept activation vectors and regions}
Note, the focus here is not so much the specific explanations produced by CAV and CAR classifiers but their ``existence" and stability, as we will investigate the fundamental properties of these methods and their interaction with our concept database.

CAVs are defined as ``the normal vector to a hyperplane separating examples without a concept and examples with a concept in the model activations" \cite{kim2018tcav}. We train CAVs as linear classifiers with stochastic gradient descent (using the hinge loss and $\ell_2$ regularization). 

CARs pose as an extension to CAVs, relaxing the assumption of linear separability of the concepts. Using the kernel trick, CARs can be determined by fitting support vector classifiers with radial basis function kernel.

For training of both CAVs and CARs, we use the code kindly provided in \cite{crabbe2022car}\footnote{\url{https://github.com/JonathanCrabbe/CARs}}. For CAV and CAR training, we use balanced data sets, i.e., the same number of positive and negative examples (usually 200 if not stated otherwise). We use $80\%$ for training and $20\%$ for testing. For testing on labeled data (i.e., Pascal VOC), we sample 200 positive and 200 negative examples.

We run two main experiments: First, we study the accuracy and robustness (\ref{sec:acc_robustness}) of CAVs and CARs, and, secondly, the alignment of concepts and sub-concepts, see (\ref{sec:alignment}). 

\subsection{Machine Learning Models}
We run all experiments with four different models: data2vec \cite{baevski2022data2vec}, the Vision Transformer (ViT) \cite{dosovitskiy2020vit}, RoBERTa \cite{liu2019roberta} and BERT \cite{devlin2018bert}.

For experiments with images, we use pretrained and fine-tuned versions of data2vec \cite{baevski2022data2vec} and ViT \cite{dosovitskiy2020vit}. Both models have been pretrained on ImageNet 21k and fine-tuned on ImageNet 1k \cite{ImageNet}. 

For experiments with text, we use pretrained and fine-tuned versions of RoBERTa \cite{liu2019roberta} and BERT \cite{devlin2018bert}. RoBERTa was fine-tuned on the ``Go Emotions" dataset \cite{demszky2020goemotions} and BERT was fine-tuned on the ``emotion" dataset \cite{saravia2018carer}. 

We retrieve all models as trained from \url{huggingface.co}, details of the exact models are shown in \autoref{tab:models}.
\begin{table}[ht]
    \centering
    %\scriptsize
    % \tabcolsep=0.11cm
    \begin{tabular}[\textwidth]{l|c|c}
       Model  & Pretrained & Fine-tuned \\
       \hline
        data2vec & facebook/data2vec-vision-base & facebook/data2vec-vision-base-ft1k \\
        ViT & google/vit-base-patch16-224-in21k & google/vit-base-patch16-224 \\
        RoBERTa & FacebookAI/roberta-base & SamLowe/roberta-base-go\_emotions \\
        BERT & google-bert/bert-base-uncased & bhadresh-savani/bert-base-uncased-emotion    \\[2mm]   
    \end{tabular}
    \caption{Overview of used models, all downloaded from \url{huggingface.co}.}
    \label{tab:models}
\end{table}
\subsection{Accuracy and robustness}
\label{sec:acc_robustness}
To measure how well-defined and robust our concept data are, we perform three experiments. We measure the test-set accuracy of CAVs and CARs and the agreement of CAVs and CARs trained on the same data. We investigate the dependence between the amount of data points or a varying negative set and the accuracy. And finally, we test how robust to distribution shifts the CAVs and CARs are.

\subsubsection{Accuracy of CAVs and CARs}
For testing the accuracy of CAVs and CARs, we use them directly as classifiers and apply 10-fold cross-validation to measure the accuracy. We measure the accuracy after each layer, it is a binary classification between the positive set (concept data) and negative set (random data). We also measure the agreement between CAVs and CARs trained on the same data, which is defined as the percentage of data that are classified into the same class. 

\subsubsection{Role of data set size}
To measure how sensitive CAVs are to the choice of negative examples, we sample the negative set ten times and calculate the cosine similarities between all the resulting CAVs. Cosine similarity is defined as $S_C(x,y) = \frac{x \cdot y}{\Vert x\Vert \Vert y\Vert}$. 

We increase the number of training data (30, 50, 200, and 1000 per concept) to explore the sensitivity to sample sizes: How much data do we need for the concepts to be well-defined?  

\subsubsection{Distribution shifts}
Finally, to explore how robust the CAVs and CARs are to distribution shifts, we test the CAVs and CARs on the additional set of labeled data (Pascal VOC images \cite{pascal-voc-2012}). Moreover, we train CAVs on the labeled data and measure the cosine similarity between CAVs trained on the same concept but from different distributions (here Wikimedia vs. Pascal VOC).

\subsection{Alignment of concepts and sub-concepts}
\label{sec:alignment}
Do self-supervised models ``understand" the relation between a concept and its sub-concepts? We hypothesize that if models develop a similar representation structure as humans, CAVs for sub-concepts should be similar to the CAV for the main concept since these are semantically close. To test this hypothesis, we explore the similarity of CAVs from concepts and their sub-concepts. For this, we choose three main concepts: \textit{sport} (Q349), \textit{fruit} (Q3314483), and \textit{motor vehicle} (Q1420). For each of the main concepts, we generate a knowledge graph of its sub-concepts with a depth of two levels (from WordNet, matched to Wikidata items). We reject sub-concepts that contain fewer than 50 images/sentences because training with such a limited amount of data would introduce noise in our experiments. In total, we have 47 sub-concepts of \textit{sport}, 23 sub-concepts of \textit{fruit} and 31 sub-concepts of \textit{motor vehicle} for text and 43 sub-concepts of \textit{sport}, 39 sub-concepts of \textit{fruit} and 27 sub-concepts of \textit{motor vehicle} for images.
We train CAVs and CARs for each of the main concepts and all sub-concepts. We use at most 200 data per concept or all that is available.

We define three different evaluation criteria to compare how similar are concepts within each group and across the groups. In all three procedures, we average the results for all inter-group and intra-group pairs. For example, the pair \textit{sport - football} contributes to the concept of sport, \textit{ orange - apple} to the concept of fruit, and \textit{ sport - orange} to the non-related concepts. If the final results for the concept groups are significantly different than the non-related group, it means that the sub-concepts are more similar if they are close in the human-defined knowledge graph.

First, we use the fact that both CAV and CAR are classifiers and test how many of the sub-concept data are classified as belonging to the main concept. We take all sub-concept data (at most 10000 per concept) and evaluate the CAV and CAR of the main concept on it.

Second, we evaluate the cosine similarity between CAVs in two ways:
\begin{enumerate}
    \item We compute the cosine similarity between CAVs and average all the intra-group pairs and inter-group pairs. 
    \item We run a triplet similarity experiment, inspired by odd-one-out comprehension tests \cite{hebart2020triplet}: we sample triplets of sub-concepts such that exactly two of them share the same main concept. We choose the pair of the triplet that has the largest cosine similarity (i.e., is the most similar). For each pair of sub-concepts, we determine the proportion of times that it was chosen as the closest pair out of all times it appeared in a triplet. We average these proportions for all inter-group pairs and all intra-group pairs.
\end{enumerate}
In both methods, we report the average result over all layers of the model.
\begin{figure}[htp]
    \centering
    \includegraphics[width=0.9\linewidth]{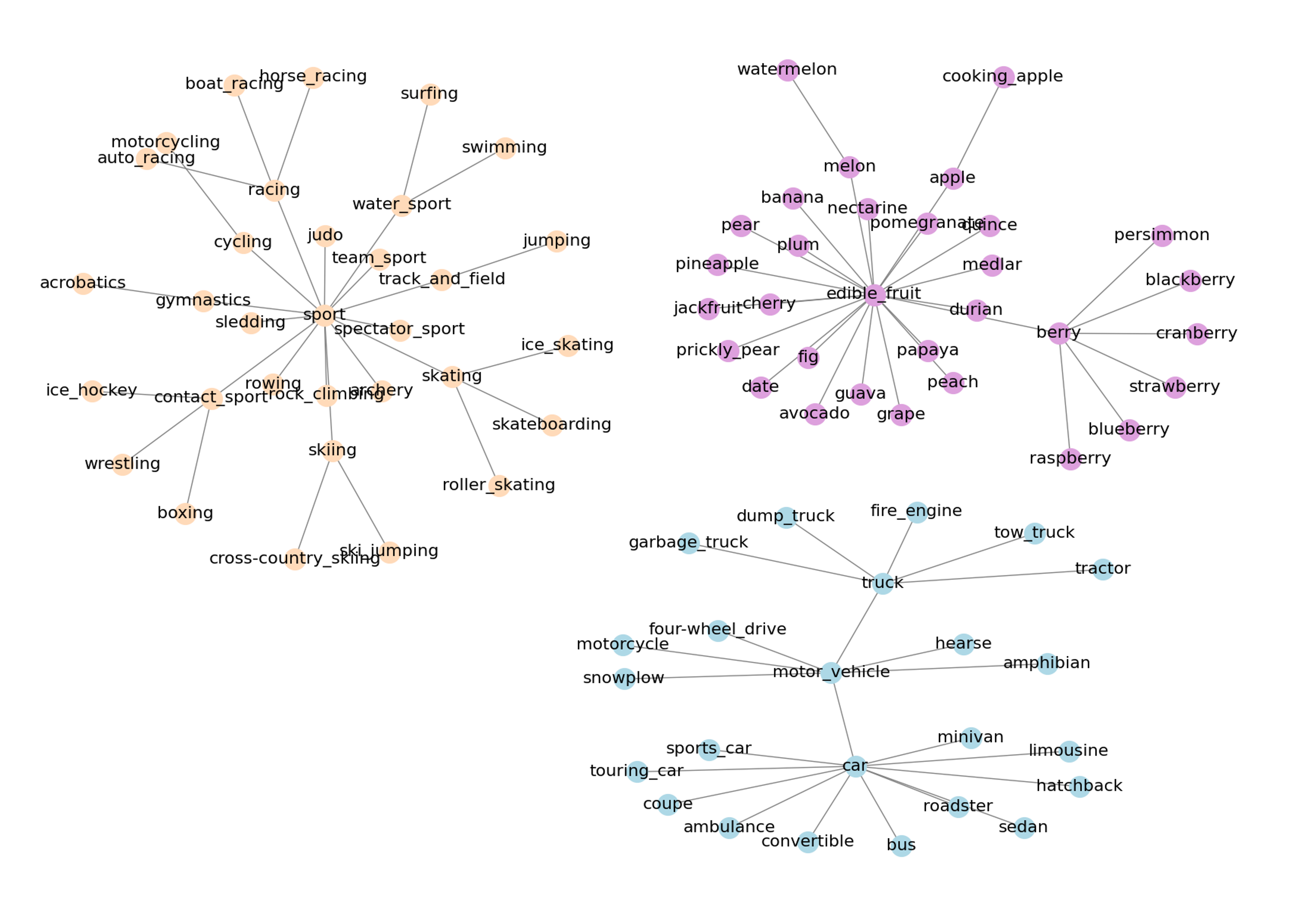}
    \caption{Extracted KGs from WordNet for the main concepts \textit{sport} (orange), \textit{fruit} (purple) and \textit{motor vehicle} (blue) -- only sub-concepts with more than 50 images available are shown. The KGs for text are slightly different since there we filter out sub-concepts with fewer than 50 sentences.}
    \label{fig:KGs}
\end{figure}
\begin{figure}[htp]
    \centering
    \begin{subfigure}[t]{0.45\linewidth}
    \includegraphics[width=\linewidth]{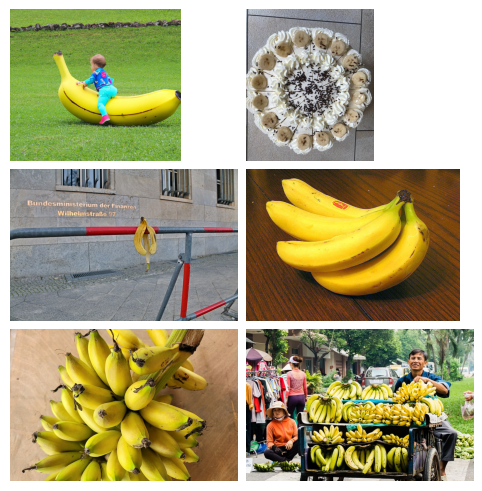}
    \caption{Banana (Q503)}
    \end{subfigure}
    \begin{subfigure}[t]{0.45\linewidth}
    \includegraphics[width=\linewidth]{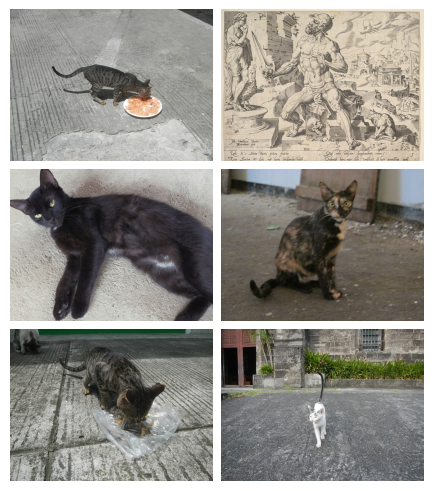}
    \caption{Cat (Q146)}
    \end{subfigure}
    \caption{Examples of images for the concept ``banana" and ``cat" from Wikimedia Commons (CC0-License).}
    \label{fig:banana}
\end{figure}
\begin{figure}[htp]
    \centering
    \includegraphics[width=\textwidth]{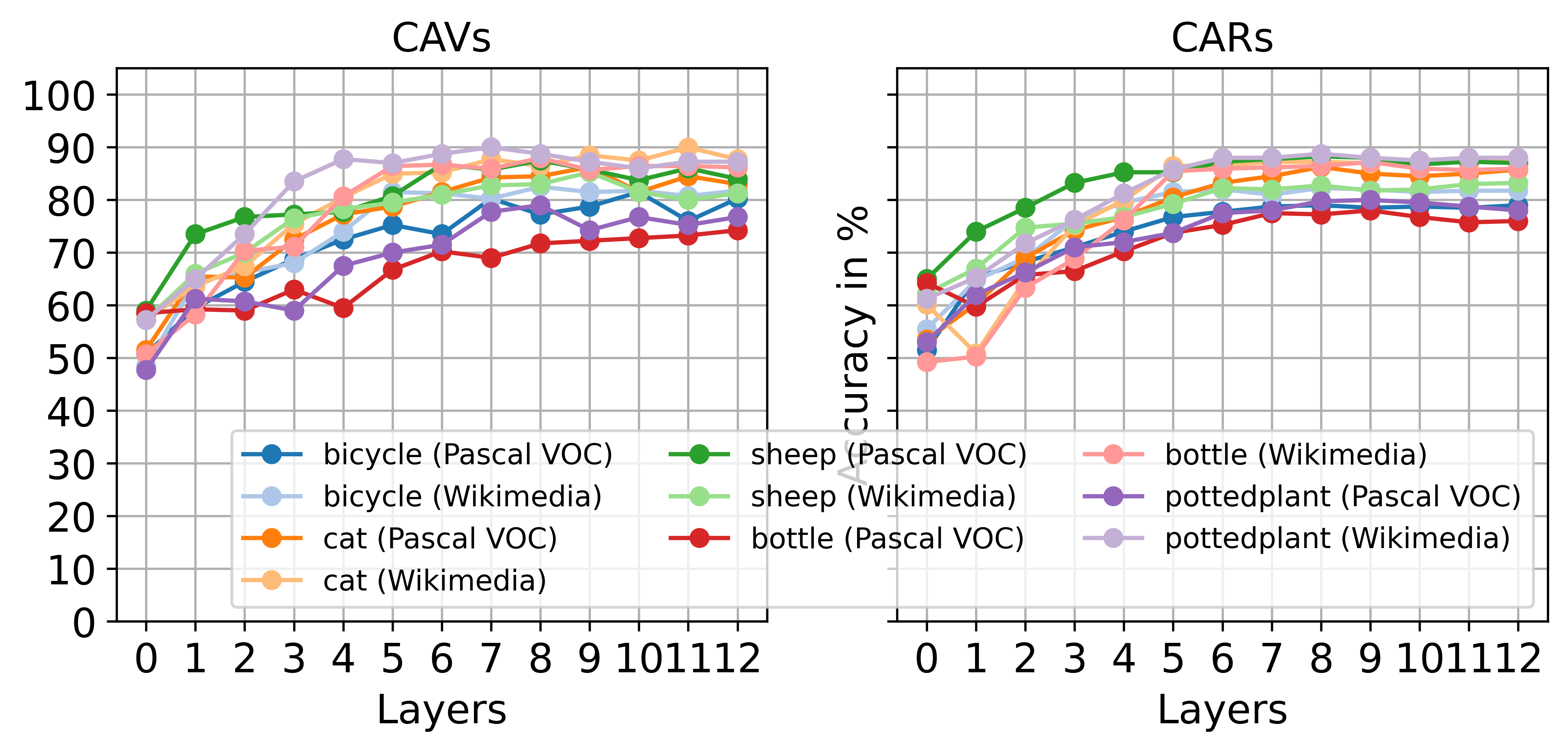}
    \caption{Accuracy of CAVs and CARs trained on both Wikimedia and Pascal VOC data with 200 concept images (using 10-fold cross-validation) for data2vec. Baseline performance achieved by random guessing amounts to 50\%.
    }
    \label{fig:acc_both}
\end{figure}
\section{Results}
\subsection{Knowledge graphs can assist the definition of data-driven concepts}
We extracted KGs with two levels of sub-concepts from WordNet for three main concepts: \textit{sport}, \textit{fruit}, and \textit{motor vehicle} and images and text for each concept and all sub-concepts. The three extracted KGs (after excluding sub-concepts with fewer than 50 available images) can be seen in \autoref{fig:KGs}. 

We compare Wikimedia and Pascal VOC images and the resulting CAVs and CARs trained on those images for the five concepts \textit{bicycle, bottle, cat, potted plant} and \textit{sheep}. Some examples of images retreived from Wikimedia can be seen in \autoref{fig:banana}. The images show a wide variety of the chosen concept.

The comparison of the accuracy of CAVs and CARs trained on images from Wikimedia and Pascal VOC (see \autoref{fig:acc_both}) shows that CAVs and CARs trained on Wikimedia have a comparable accuracy on the test set, and even outperforming the CAVs and CARs trained on the Pascal VOC dataset in 4 of 5 concepts. The relatively high accuracy of the CAVs and CARs indicates well-defined concepts and concept data. Important to note here, is the general challenge that concepts can be harder to learn than the general classes they should explain as pointed out by \cite{ramaswamy2023overlooked}, so the accuracy of CAVs and CARs should always be tested and concepts with low accuracy need to be redefined or removed to ensure reliable explanations. In early layers the accuracy of CAVs and CARs is generally lower, so explanations based on these should be evaluated critically. 

The great variety of images sourced from Wikimedia and the better\--per\-for\-ming classifiers based on these images indicate the great potential of the use of Wikimedia as a data source and the importance of a diverse dataset. This can lead to better-defined CAVs and CARs which in a downstream explanation task can foster robustness of explanations and, therefore, increase trust in the provided explanation.

\subsection{Robustness of knowledge graph derived concepts}
We test the robustness of CAVs and CARs trained on the KG-based concepts to the variation of the negative set and the size of the training set. A varying negative set should not influence the performance and definition (i.e., direction) of CAVs and CARs. The cosine similarities for the three concepts \textit{sport}, \textit{fruit} and \textit{motor vehicle} can be seen in \autoref{tab:sanity}, for all models and concepts a high similarity is observed. The results are evidence of high robustness to varying negative sets and underline that concepts are well-defined. 
\begin{table}[htp]
  \centering
  \begin{tabular}{|l||c|c|c|} \hline
     Model                      & sport & fruit & motor vehicle  \\ \hline \hline
     RoBERTa (text) 	 & 0.72 $\pm$ 0.02 	 &   0.80 $\pm$ 0.01 	 &  0.86 $\pm$ 0.01 	 \\
     BERT (text) 	 & 0.59 $\pm$ 0.02 	 &  0.73 $\pm$ 0.01 	 &  0.76 $\pm$ 0.01 \\ \hline
     data2vec (images) 	 & 0.71 $\pm$ 0.03 	 &  0.75 $\pm$ 0.02 	 &  0.75 $\pm$ 0.03 	 \\
     ViT (images)   & 0.62 $\pm$ 0.02 	 &  0.67 $\pm$ 0.02 	 &  0.66 $\pm$ 0.03 \\ \hline
  \end{tabular}
  \caption{Averages and standard error of the means for cosine similarities for various random negative sets (10 repetitions) for Wikimedia images and Wikipedia sentences averaged over all the layers for pretrained models (results for fine-tuned models are almost identical).}
  \label{tab:sanity}
\end{table}
\begin{figure}[htp]
    \centering
    \includegraphics[width=\textwidth]{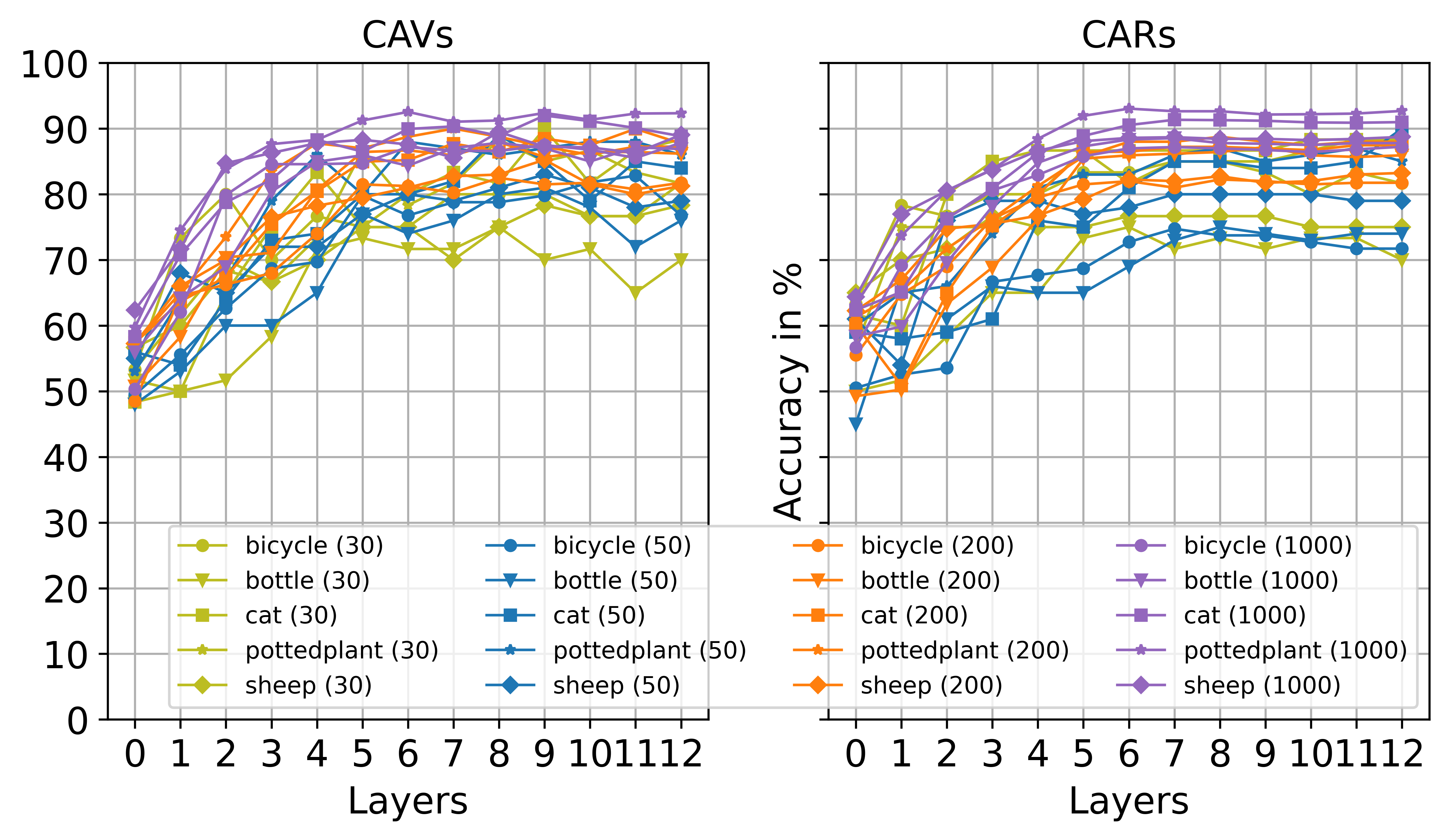}
    \caption{Accuracy of CAVs and CARs depending on the number of training data (for positive class). The lines show the means of accuracies over 10-fold cross-validation. In both cases, more training data leads to higher accuracy. Results shown for data2vec, similar trends are observed for all models.}
    \label{fig:size_dependence}
\end{figure}
\begin{figure}[htp]
    \centering
    \includegraphics[width=\textwidth]{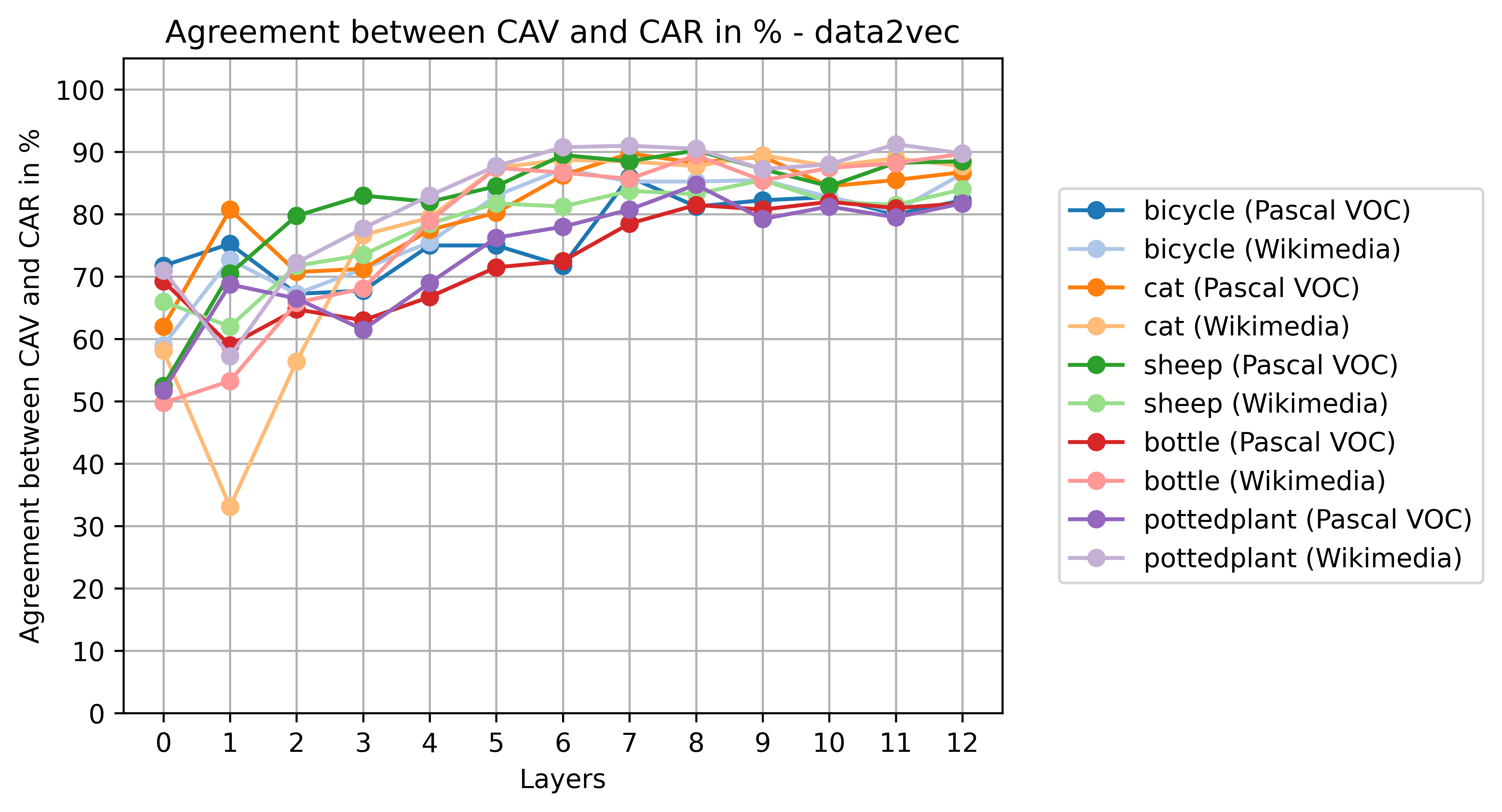}
    \caption{Agreement between CAVs and CARs: percentage of predictions that were the same for corresponding CAV and CAR. Results shown for data2vec.}
    \label{fig:agreement}
\end{figure}
Performance increases with the size of the training set as shown in \autoref{fig:size_dependence}. Especially for low numbers of training samples (30, 50), the accuracy is relatively low and unstable across layers for some concepts. For a higher number of training samples ($>200$), the accuracy improves and stabilizes across layers. The highest accuracy is observed for 1000 data samples. This demonstrates the need for a diverse and large enough training set, which the KG-based approach can provide. Once more, we notice reduced accuracy in the initial layers of the network, emphasizing the need for caution when interpreting the explanations for these early layers, regardless of the quantity of training data available.
\begin{figure}[htp]
    \centering
    \begin{subfigure}[t]{0.9\textwidth}
        \includegraphics[width=\textwidth]{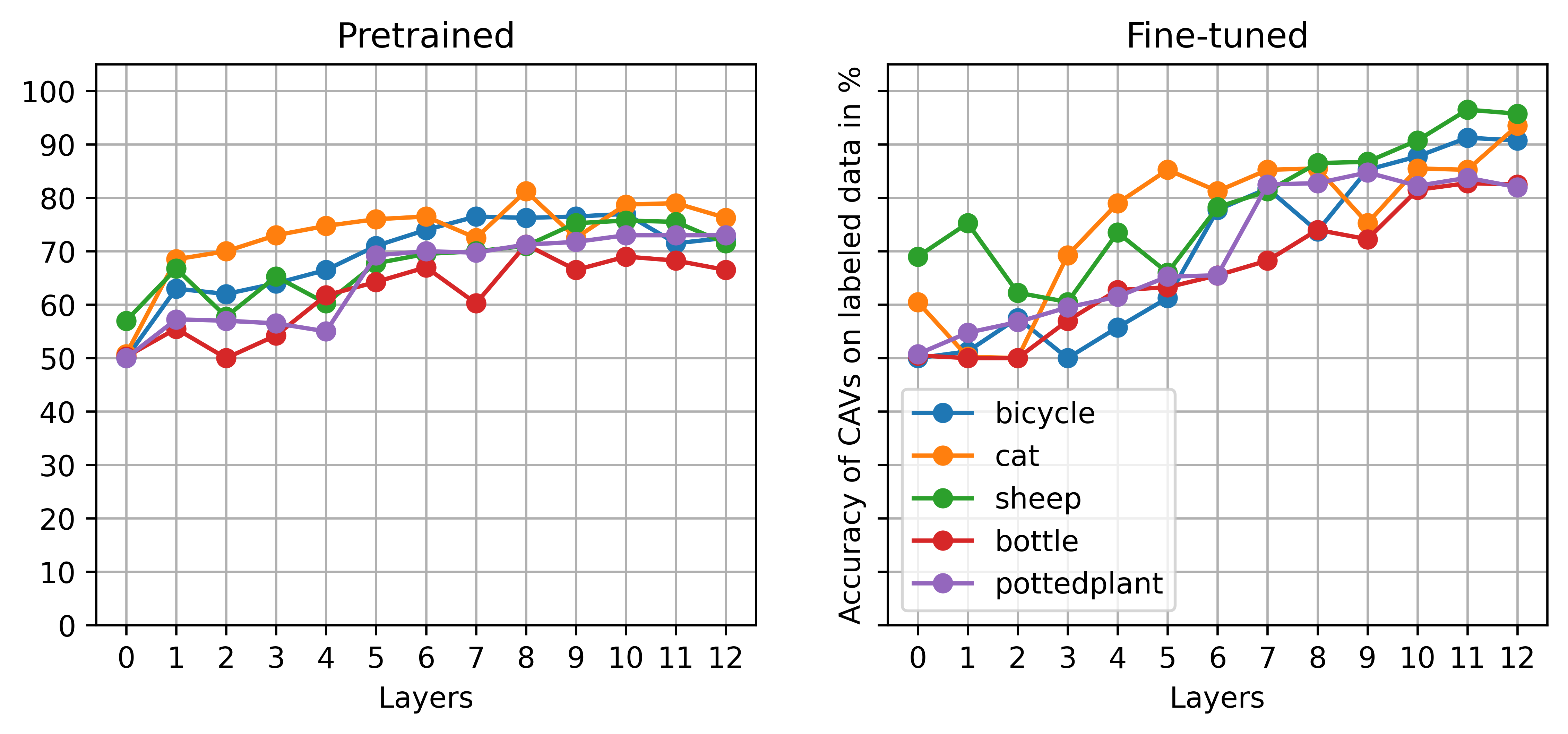}
        \caption{Accuracy of CAVs trained on Wikimedia data and tested on Pascal VOC.}    
        \label{fig:ood_CAV}
    \end{subfigure}
    \begin{subfigure}[t]{0.9\textwidth}
        \includegraphics[width=\textwidth]{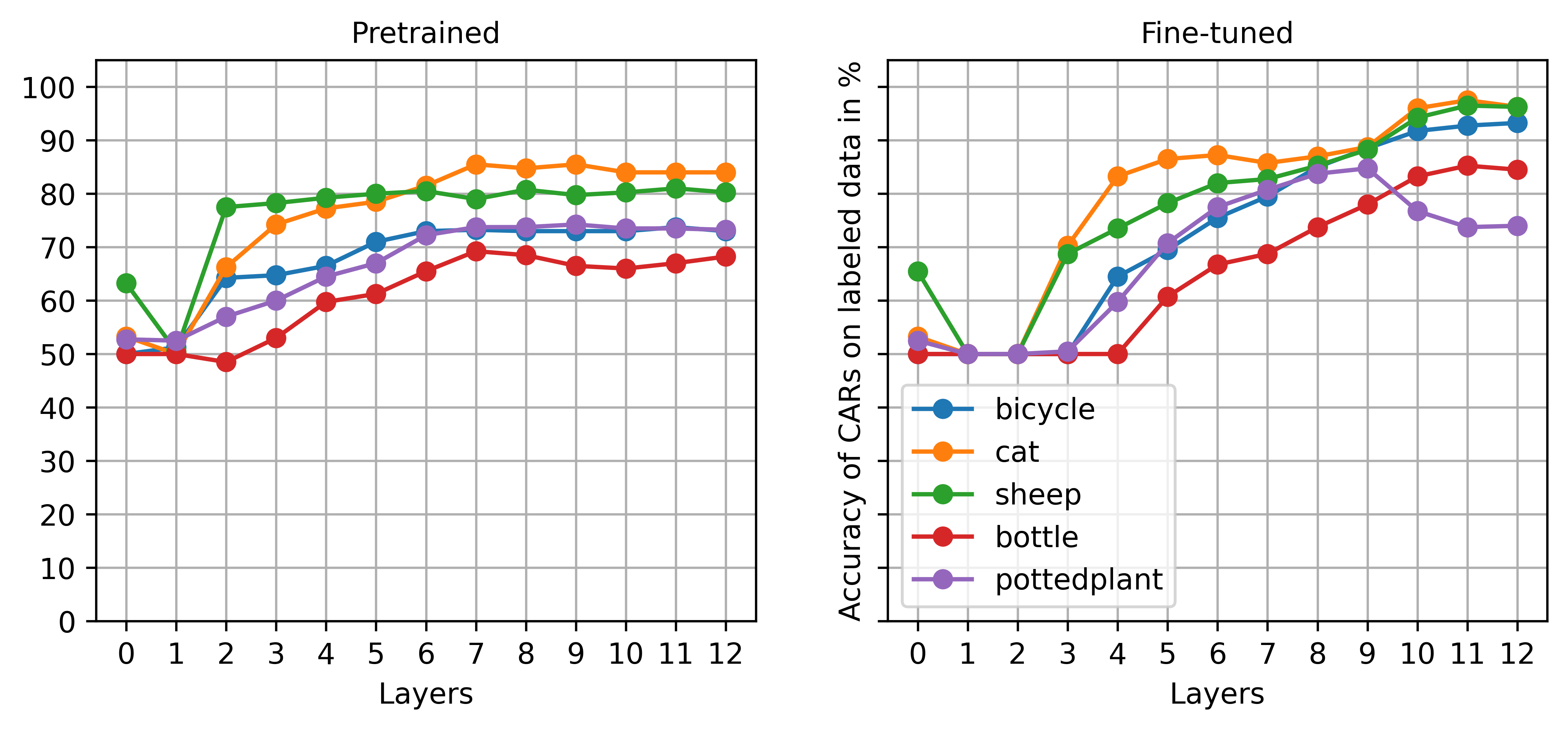}
        \caption{Accuracy of CARs trained on Wikimedia data and tested on Pascal VOC.}    
        \label{fig:ood_CAR}
    \end{subfigure}
    \begin{subfigure}[t]{0.9\textwidth}
        \includegraphics[width=\textwidth]{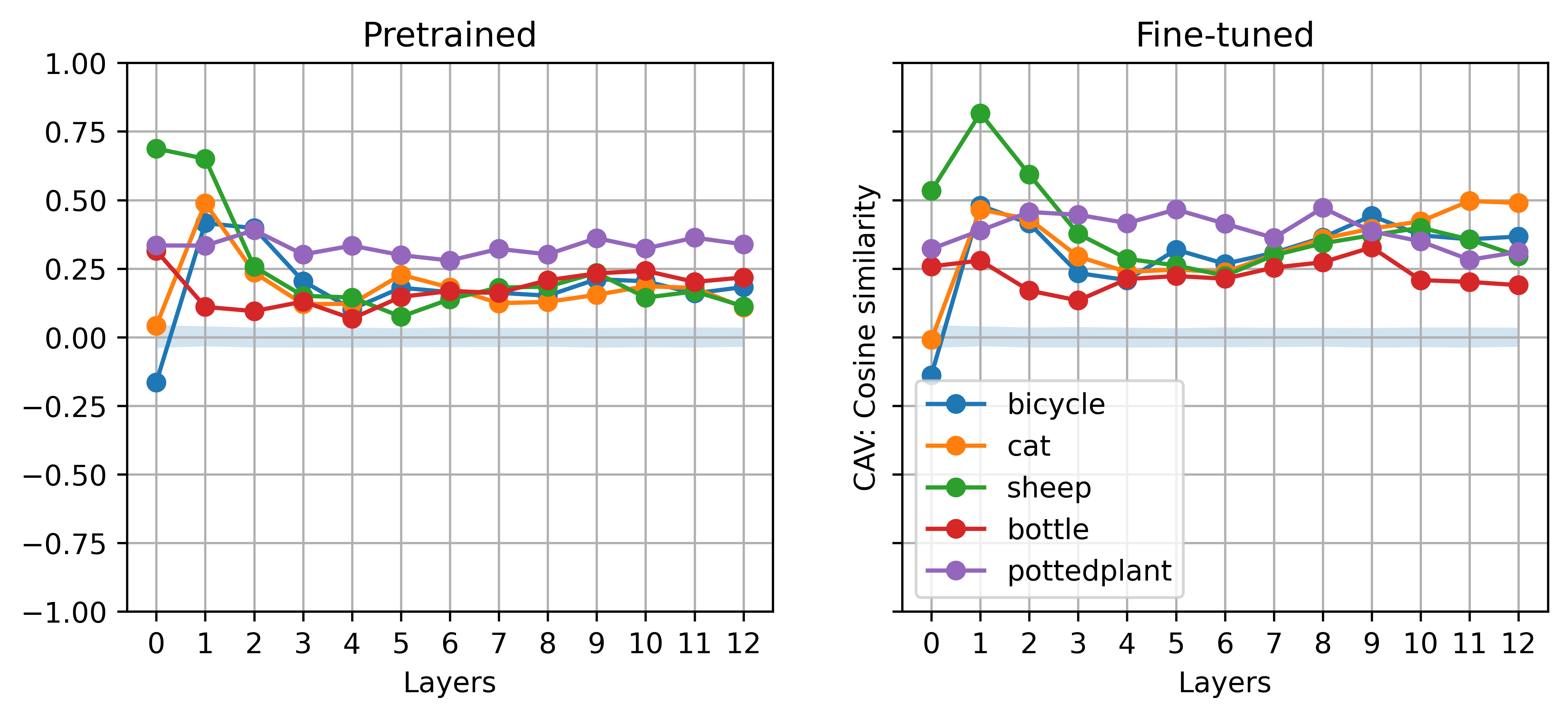}
        \caption{Cosine similarity between CAVs trained on Wikimedia data, and on Pascal VOC. The light blue stripes are baselines computed by permutation test on the CAVs (100 repetitions).}    
        \label{fig:ood_cosine}
    \end{subfigure}
    \caption{Evaluation of CAVs and CARs trained on Wikimedia and tested on Pascal VOC (for data2vec).}
    \label{fig:ood}
\end{figure}
The results for all experiments are very similar when comparing CAVs and CARs. CAVs and CARs also agree on more than $80\%$ of the predictions for concepts defined by Wikimedia (see \autoref{fig:agreement}) in later layers.
The high agreement between CAVs and CARs is evidence that concept-based XAI is robust towards the choice of CAVs vs.\ CARs.

We finally test out-of-distribution (OOD) performance by comparing CAVs and CARs trained on Wikimedia and applied to images from Pascal VOC (see \autoref{fig:ood_CAV} and \autoref{fig:ood_CAR}). We also compare CAVs trained on Wikimedia with CAVs trained on Pascal VOC via cosine similarity (see \autoref{fig:ood_cosine}). We observe accuracies of 65-85\% for pretrained models and even higher accuracies of 73-97\% for fine-tuned models in later layers (compared to 80-90\% when tested on Wikimedia), indicating the robustness of the CAVs and CARs to OOD data. CAVs trained on different datasets have a cosine similarity of 0.15-0.35 for pretrained models and 0.2-0.5 for fine-tuned models (compared to 0.7-0.85 for varying negative sets). This shows some degree of alignment between the CAVs and CARs trained on different databases, however, the variation between them should not be disregarded. The dependence of explanations on the dataset has also been pointed out before \cite{ramaswamy2023overlooked} and highlights the need for well-defined and well-aligned concept datasets that meet the user's expectations and intentions.
\subsection{Alignment of Human and Machine representations}
\begin{figure}[htp]
    \centering
    \includegraphics[width=0.9\textwidth]{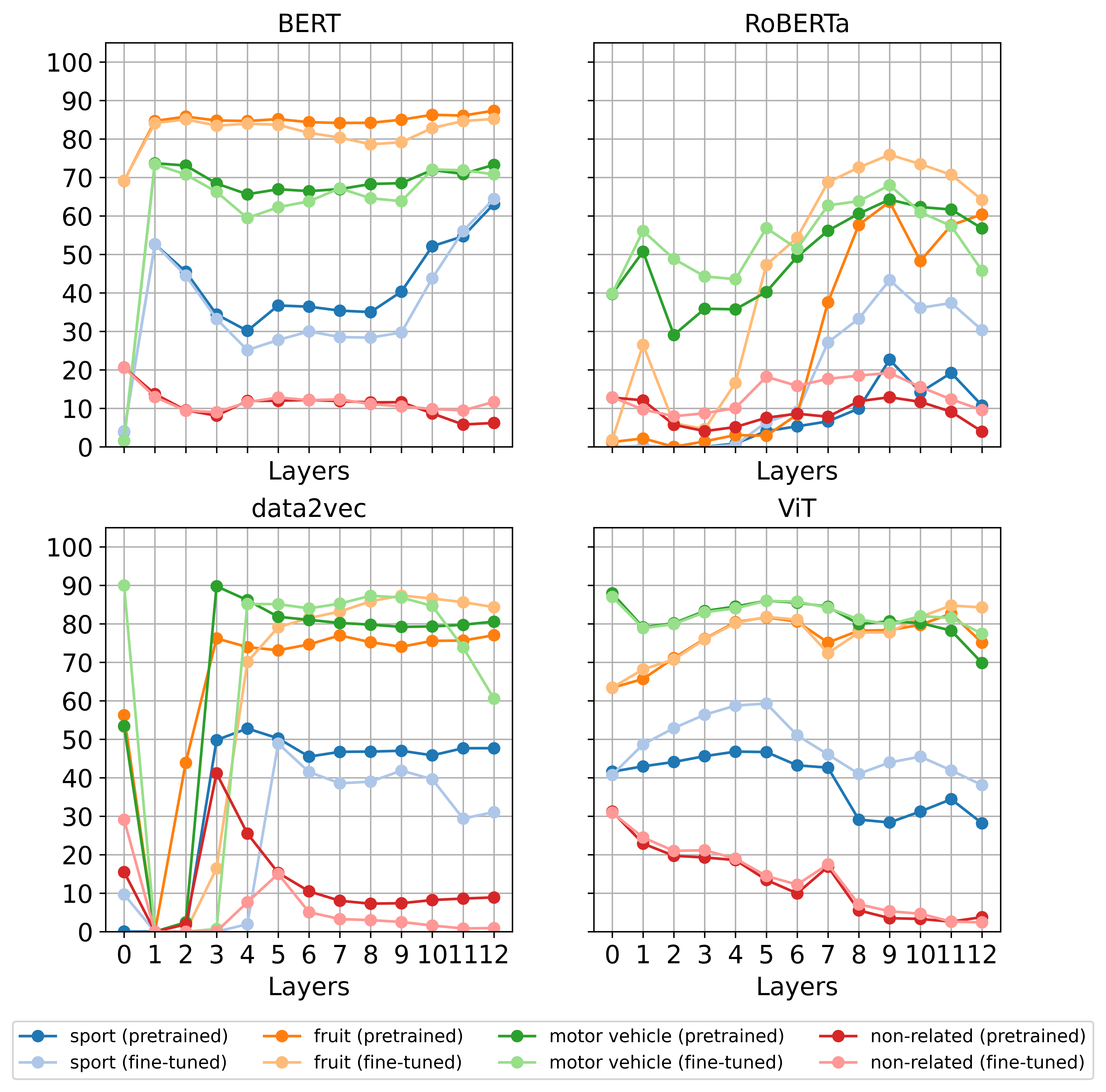}
    \caption{Percentage of sub-concept data classified by CARs as belonging to the main concept.}
    \label{fig:subconcepts_accuracy}
\end{figure}

Lastly, we test the alignment between the representations of the model and humans. We measure the similarity between the KG structure and CAV directions and find that related concepts and sub-concepts are more aligned than non-related concepts. Related concepts show very high agreement on the triplet experiment, around 90\% for \textit{fruit} and \textit{motor vehicle} and around 60\% for \textit{sport} compared to around 10\% for non-related concepts for all models. The CAVs also show higher cosine similarity for related concepts vs. non-related concepts. The detailed values for triplets and cosine similarity can be found in \autoref{tab:subconcepts_cosine_similarity}. Note that the cosine similarities, in general, should not be expected to be higher than in \autoref{tab:sanity}, where we measured the agreement for varying negative sets. As we see, the triplets experiment has a higher signal-to-noise ratio. The table suggests alignment of the internal model representations with a human-defined structure (i.e., a KG), which is crucial for successful communication between the two \cite{gardenfors2000conceptual}.
\begin{table}[htp]
  \centering
  \begin{subtable}{\linewidth}
  \centering
  \begin{tabular}{|l||l|l|l|l|} \hline
     Model & Sport    & Fruit    & Motor vehicle   & Non-related \\ \hline \hline
     RoBERTa (text) & 0.334 $\pm$ 0.004 	 &  0.512$\pm$ 0.010 	 &  0.515$\pm$ 0.006 	 &  0.213$\pm$ 0.002 	 \\
     RoBERTa fine-tuned (text)  & 0.276 $\pm$ 0.012 	 &  0.451$\pm$ 0.006 	 &  0.439$\pm$ 0.011 	 &  0.179$\pm$ 0.005 	 \\
     BERT (text)    & 0.253 $\pm$ 0.007 	 &  0.456$\pm$ 0.015 	 &  0.431$\pm$ 0.011 	 &  0.152$\pm$ 0.006 	 \\
     BERT fine-tuned (text) & 0.238 $\pm$ 0.006 	 &  0.436$\pm$ 0.013 	 &  0.407$\pm$ 0.011 	 &  0.144$\pm$ 0.006 	 \\ \hline
     data2vec (images)  & 0.222 $\pm$ 0.011 	 &  0.391$\pm$ 0.016 	 &  0.428$\pm$ 0.034 	 &  0.092$\pm$ 0.019 	 \\ 
     data2vec fine-tuned (images)   & 0.216 $\pm$ 0.021 	 &  0.435$\pm$ 0.023 	 &  0.435$\pm$ 0.053 	 &  0.097$\pm$ 0.020 	 \\ 
     ViT (images)   & 0.153 $\pm$ 0.014 	 &  0.277$\pm$ 0.028 	 &  0.340$\pm$ 0.032 	 &  0.089$\pm$ 0.009 	 \\ 
     ViT fine-tuned (images)    & 0.153 $\pm$ 0.013 	 &  0.290$\pm$ 0.028 	 &  0.318$\pm$ 0.030 	 &  0.088$\pm$ 0.010 	 \\ \hline
    \end{tabular}
  \caption{Averages of cosine similarities}
  \end{subtable}
  \begin{subtable}{\linewidth}    
  \centering
  \begin{tabular}{|l||l|l|l|l|} \hline
     Model & Sport    & Fruit    & Motor vehicle   & Non-related \\ \hline \hline
     RoBERTa (text) & 0.631 $\pm$ 0.008 	 &  0.918$\pm$ 0.001 	 &  0.937$\pm$ 0.007 	 &  0.101$\pm$ 0.003 	 \\
     RoBERTa fine-tuned (text)  & 0.600 $\pm$ 0.010 	 &  0.931$\pm$ 0.004 	 &  0.925$\pm$ 0.008 	 &  0.110$\pm$ 0.003 	 \\
     BERT (text)   & 0.646 $\pm$ 0.010 	 &  0.917$\pm$ 0.010 	 &  0.943$\pm$ 0.010 	 &  0.102$\pm$ 0.005 	 \\ 
     BERT fine-tuned (text) & 0.632 $\pm$ 0.008 	 &  0.920$\pm$ 0.010 	 &  0.937$\pm$ 0.010 	 &  0.107$\pm$ 0.004 	 \\ \hline
     data2vec (images) & 0.630 $\pm$ 0.025 	 &  0.954$\pm$ 0.009 	 &  0.924$\pm$ 0.037 	 &  0.094$\pm$ 0.010 	 \\ 
     data2vec fine-tuned (images)   & 0.589 $\pm$ 0.025 	 &  0.968$\pm$ 0.009 	 &  0.903$\pm$ 0.035 	 &  0.102$\pm$ 0.009 	 \\ 
     ViT (images)   & 0.552 $\pm$ 0.023 	 &  0.872$\pm$ 0.046 	 &  0.913$\pm$ 0.048 	 &  0.126$\pm$ 0.017 	 \\
     ViT fine-tuned (images)    & 0.561 $\pm$ 0.021 	 &  0.888$\pm$ 0.048 	 &  0.903$\pm$ 0.048 	 &  0.123$\pm$ 0.018 	 \\ \hline
  \end{tabular}
  \caption{The triplet experiment}
  \end{subtable}  
  \caption{Alignment between human and machine representations -- similarity of the sub-concepts belonging to the same main concept (\textit{sport}, \textit{fruit} and \textit{motor vehicle}), and of the unrelated concepts. See \autoref{sec:alignment} for more details. The uncertainties are the standard error of the mean. The results show that concepts that are close to each other in knowledge graphs (i.e., human representations) are also close to each other in the machine representations.}
  \label{tab:subconcepts_cosine_similarity}
\end{table}
When evaluating CARs trained on the main concepts and tested on their sub-concepts (\autoref{fig:subconcepts_accuracy}), it becomes evident that sub-concepts are often classified as their main concept (and therefore being part of that concept) in later layers. This aligns with earlier observations that data belonging to the same class/concept form more convex regions in later layers of the network \cite{tetkova2023convex} resulting in higher performance of the classifiers. The results of the same experiment with CAVs are very noisy, hugely varying from layer to layer. This suggests that CAVs are more stable as directions than as classifiers and CARs are better at depicting the semantic structure of the representations.
The difference between the concepts in all experiments in this section can potentially be explained by their KG structure. The concept \textit{sport} and its sub-concepts seem to be less aligned than the two concepts \textit{fruit} and \textit{motor vehicle}. When examining the structure of their KGs (\autoref{fig:KGs}), it becomes apparent that \textit{sport} is the most branched out KG with half of the first-level sub-concepts containing several second-level sub-concepts, while the two other KGs only have two or three first-level sub-concepts containing second-level sub-concepts. This means that \textit{sport} is a much broader concept than the other two, which could explain why the sub-concepts of sport are more spread out in the model representations and have a lower degree of alignment with their main concept. This again underlines the importance of well-defined concepts, that are aligned with the user's intention, as the explanations can differ depending on the granularity of the concept definition. 
\section{Conclusion}
We introduce an interactive workflow for concept definition and automatic data retrieval based on knowledge graphs. Using publicly available resources, such as the Wikimedia project, we are able to create larger concept databases than available labeled databases (such as Pascal VOC) with minimal supervision, which lead to comparable or even better accuracies for concept activation vectors/regions (CAVs and CARs). We generally observe lower accuracy and agreement of CAVs and CARs in the early layers of all networks, suggesting that explanations derived from the early layers should be viewed critically. We show that explanations based on the retrieved concept databases are robust to in-distribution shifts like variations in the negative set. %For fine-tuned models 
We even find a certain degree of robustness in later layers to out-of-distribution shifts (i.e., using a different dataset). However, it is important to keep in mind that different concept datasets can lead to different CAVs and CARs and therefore varying explanations \cite{ramaswamy2023overlooked}. This highlights the importance of aligning the concept definition and database with the intention of the user, as the explanation can depend strongly on the context of the concept. Finally, we show that networks learn a similar relation of concepts to sub-concepts as in human generated knowledge graphs, suggesting some inherent alignment. This human-machine alignment is essential for successful communication and underscores the promising future of concept-based explainability. 
\section{Acknowledgements}
This work was supported by the DIREC Bridge project Deep Learning and Automation of Imaging-Based Quality of Seeds and Grains, Innovation Fund Denmark grant number 9142-00001B. This work was supported by the Pioneer Centre for AI, DNRF grant number P1 and the Novo Nordisk Foundation grant NNF22OC0076907 ”Cognitive spaces - Next generation explainability”.
\bibliographystyle{splncs04}
\bibliography{bibliography}
\end{document}